\documentclass[11pt,a4paper]{article}
\usepackage[hyperref]{emnlp2018}
\usepackage{times}
\usepackage{latexsym}

\usepackage{url}
\usepackage{booktabs}
\usepackage{graphicx}
\usepackage{csquotes}

\aclfinalcopy % Uncomment this line for the final submission

\title{Aspect and Opinion Term Extraction for Hotel Reviews using Transfer Learning and Auxiliary Labels}

\author{Yosef Ardhito Winatmoko \\
  Jheronimus Academy of Data Science \\
  's-Hertogenbosch, The Netherlands \\
  {\tt y.a.winatmoko@uvt.nl} \And
  Ali Akbar Septiandri \\
  Universitas Al Azhar Indonesia \\
  Jakarta, Indonesia \\
  {\tt aliakbar@if.uai.ac.id} \AND
  Arie Pratama Sutiono \\
  Ninja Van \\
  Singapore \\
  {\tt arie.pratama.s@gmail.com} \\}

\date{}

\begin{document}
\maketitle
\begin{abstract}
Aspect and opinion term extraction is a critical step in Aspect-Based Sentiment Analysis (ABSA). Our study focuses on evaluating transfer learning using pre-trained BERT \citep{devlin2018bert} to classify tokens from hotel reviews in bahasa Indonesia. The primary challenge is the language informality of the review texts. By utilizing transfer learning from a multilingual model, we achieved up to 2\% difference on token level F1-score compared to the state-of-the-art Bi-LSTM model with fewer training epochs (3 vs. 200 epochs). The fine-tuned model clearly outperforms the Bi-LSTM model on the entity level. Furthermore, we propose a method to include CRF with auxiliary labels as an output layer for the BERT-based models. The CRF addition further improves the F1-score for both token and entity level.
\end{abstract}

\section{Introduction}

Sentiment analysis \citep{pang2008opinion} in review text usually consists of multiple steps. In this study, we focus on the aspect and opinion term extraction from the reviews for ABSA \citep{liu2012survey}. While some work has been done in this task \citep{wang2017coupled, Fernando2019, xue2018aspect}, we have not seen a transfer learning approach \citep{ruder2019neural} employed for ABSA in other languages than English. Using transfer learning is especially helpful for low-resource languages \citep{kocmi2018trivial}, such as bahasa Indonesia. 

As an illustration of aspect and sentiment extraction, here is an example of a review: 

\begin{displayquote}
``Excellent location to the Tower of London. The room was a typical hotel room in need of a refresh, however clean. The staff couldn't have been more professional.''
\end{displayquote} 

In this review, some of the aspect terms are ``location'', ``hotel room'', and ``staff''. On the other hand, the corresponding sentiment terms are ``excellent'', ``typical'', ``clean'', and ``professional''.

Our main contribution in this study is evaluating BERT \citep{devlin2018bert} as a pretrained transformer model on this token classification task on hotel reviews in bahasa Indonesia. We also found that applying Conditional Random Field (CRF) \citep{lafferty2001conditional} as an output layer for BERT is not straightforward due to the subword tokenization imposed by BERT model. We propose using auxiliary labels for the special tokens to cater to the subword tokenization.

In the following sections, we describe the transfer learning approach and the auxiliary labels for the CRF. Subsequently, we elaborate on the experimental setup and the results compared with a baseline and Bi-LSTM model by \cite{Fernando2019}. Finally, we discuss the results in terms of performance and resource required for each model.

\section{Model}

\begin{figure*}[htbp]
    \centering
    \includegraphics[width=0.9\textwidth]{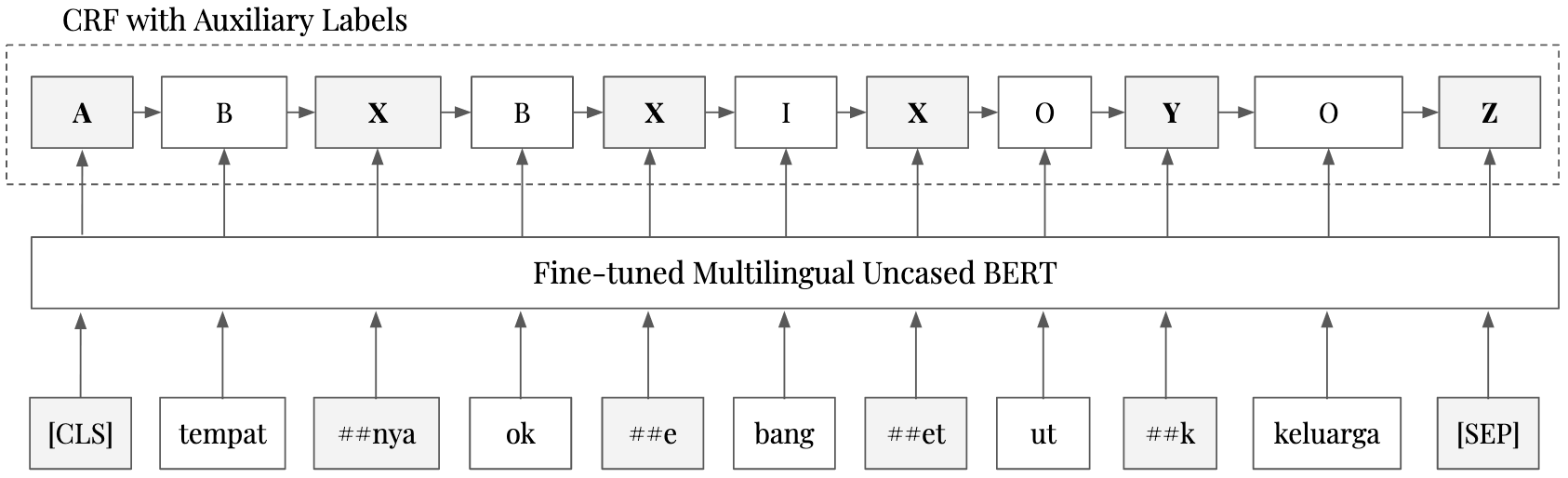}
    \caption{An example of CRF with auxiliary labels to handle special tokens and subwords tokenization imposed by the pretrained BERT model. The trailing tags (SENTIMENT and ASPECT) are omitted from the illustration.}
    \label{fig:crf}
\end{figure*}

For the transfer learning, we used the pretrained BERT-base, multilingual uncased \citep{devlin2018bert} implementation in PyTorch by \citet{HuggingFace2019}\footnote{\url{https://github.com/huggingface/transformers}}. This model is trained on 102 most popular languages in Wikipedia, including bahasa Indonesia. In the vanilla version, the output layer for BERT is based on argmax. For entity recognition tasks, CRF is commonly used to ensure that the predicted labels follow the BIO constraint \citep{lample2016neural}. However, BERT tokenizes words into subwords as inputs, which led to auxiliary labels, as shown in \autoref{fig:crf}.

We introduce four auxiliary labels to handle the special tokens: A, Z, X, and Y. A and Z correspond to the \texttt{[CLS]} and \texttt{[SEP]}, respectively. The former is a special token of BERT tokenizer to indicate the beginning of a sentence while the latter is the sentence separator token. 

X is an auxiliary label for subwords belonging to an aspect or sentiment. For instance, the word ``tempatnya'' (\textit{the place}) is tokenized into two subwords: ``tempat'' and ``\#\#nya''. We keep the original label for the first subword ``tempat'', and designate X-ASPECT as the label for any trailing subwords, in this case ``\#\#nya''. Similarly for ``oke'' (\textit{OK}) and ``banget'' (\textit{very}), which are labelled as B-SENTIMENT and I-SENTIMENT, respectively, we use X-SENTIMENT as the subword labels.

We allocate Y for any trailing subwords that are neither part of aspect nor sentiment. In the example shown in \autoref{fig:crf}, the word ``utk'' (\textit{for}) is split into two: ``ut'' and ``\#\#k''. The first subword, ``ut'' is labeled with the original label O, while ``\#\#k'' acquire the auxiliary label Y. The rest of the CRF implementation is unchanged. We adopt the CRF layer for PyTorch as implemented in the pytorch-crf library\footnote{\url{https://github.com/kmkurn/pytorch-crf}}.

\section{Experiment}

\subsection{Dataset}

We use tokenized and annotated hotel reviews on Airy Rooms\footnote{\url{https://www.airyrooms.com/}} provided by \citet{Fernando2019}\footnote{\url{https://github.com/jordhy97/final_project}}. The dataset consists of 5000 reviews in bahasa Indonesia. The dataset contains training and test sets of 4000 and 1000 reviews, respectively. The label distribution of the tokens in the BIO scheme can be seen in Table~\ref{tab:label}. Moreover, we also see this case as on the entity level, i.e., ASPECT, SENTIMENT, and OTHER labels. 

We split the training set into 3000 for the training and 1000 for the validation set to tune the hyperparameters. We found that there are 1643 and 809 unique tokens in the training and test sets, respectively. Moreover, 75.4\% of the unique tokens in the test set can be found in the training set.

\begin{table}[htbp]
    \centering
    \begin{tabular}{lrr}
    \toprule
    Label & Train & Test \\
    \midrule
    B-ASPECT 	& 7005  &	1758 \\
    I-ASPECT 	& 2292  &	 584 \\
    B-SENTIMENT & 9646  &	2384 \\
    I-SENTIMENT & 4265  &	1067 \\
    OTHER       & 39897 &	9706 \\
    \midrule
    Total       & 63105 &	15499 \\
    \bottomrule
    \end{tabular}
    \caption{Label distribution}
    \label{tab:label}
\end{table}

\subsection{Setup}

The following hyperparameters are the same for both BERT and BERT+CRF. For the \textit{learning rate}, we experimented with range $10^{-6}$ to $10^{-2}$ and in logarithmic scale. We found $10^{-4}$ as the optimal value. We employed \textit{AdamW} with the optimal learning rate as the peak value and \textit{weight decay} of $10^{-2}$ \citep{loshchilov2017decoupled}. The value for the \textit{warmup steps} is set to half of the total steps. For the \textit{batch size}, we tried 16 and 32. We found 32 to be better and used the same value for all models.

For the \textit{number of epochs}, we have different values for BERT and BERT+CRF. BERT was trained only with two epochs since it starts to overfit immediately. For BERT+CRF, we use three epochs for the token level and four for the entity level. As a baseline, a simple argmax method is employed. In the argmax method, we classify a token as the most probable label (the highest proportion) for that particular token according to the distribution in the training set.

For the evaluation metric, we use $F_1$-score because of the tag imbalance. The $F_1$-scores for the test set are based on the model trained on the 3000 training set sentences.

\subsection{Results}

The results from our experiments are summarized in \autoref{tab:result1} for token level (with BIO scheme) and \autoref{tab:result2} for entity level (without BIO). BERT corresponds to the vanilla BERT model with argmax as the output layer in the tables, while BERT+CRF utilizes CRF with auxiliary labels.

\begin{table*}[htbp]
    \centering
    \begin{tabular}{lccccc}
    \toprule
    Method & B-ASPECT & I-ASPECT & B-SENTIMENT & I-SENTIMENT & OTHER \\
    \midrule
    argmax                  & 0.777 & 0.592 & 0.810 & 0.391 & 0.851 \\
    \citet{Fernando2019}    & 0.916 & \textbf{0.873} & \textbf{0.939} & 0.886 & \textbf{0.957} \\
    BERT                    & 0.916 & 0.863 & 0.932 & 0.862 & 0.952 \\
    BERT+CRF                & \textbf{0.924} & 0.862 & 0.938 & \textbf{0.887} & 0.954 \\
    \bottomrule
    \end{tabular}
    \caption{BIO scheme (token level) $F_1$ test scores}
    \label{tab:result1}
\end{table*}

\begin{table}[htbp]
    \centering
    \begin{tabular}{lccc}
    \toprule
    Method & Aspect & Sentiment \\
    \midrule
    argmax                  & 0.81 & 0.85 \\
    \citet{Fernando2019}    & 0.89 & 0.91 \\
    BERT                    & 0.91 & 0.92 \\
    BERT+CRF                & \textbf{0.92} & \textbf{0.93} \\
    \bottomrule
    \end{tabular}
    \caption{Aspect and sentiment (entity level) $F_1$ test scores}
    \label{tab:result2}
\end{table}

\subsection{Discussion}

Based on \autoref{tab:result1}, using pretrained multilingual BERT can help achieve almost the same performance to the Bi-LSTM model \citep{Fernando2019}. The former's advantage is the required number of epochs: pretrained models needed at maximum four epochs to train, while the Bi-LSTM model was trained for 200 epochs. Furthermore, we can see that the CRF with auxiliary labels improves the $F_1$ slightly. The BERT+CRF performance is almost identical to the Bi-LSTM model. On the entity level (\autoref{tab:result2}), the BERT-based models outperform the Bi-LSTM model for both aspect and sentiment detection.

\begin{figure}[htbp]
    \centering
    \includegraphics[width=\columnwidth]{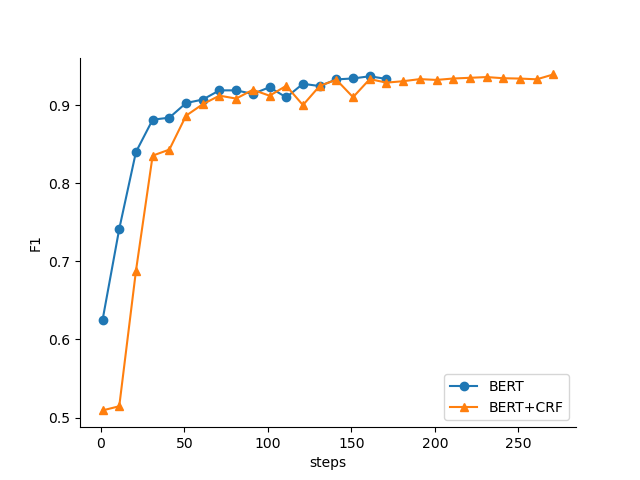}
    \caption{Token level $F_1$ for validation set}
    \label{fig:f1}
\end{figure}

\autoref{fig:f1} shows the validation $F_1$ throughout training steps (mini-batches). We can see that BERT+CRF needed more steps to reach the highest F1 compared to vanilla BERT. The $F_1$ scores tend to plateau as well, showing an indication that the models are robust and not easily overfitting.

Without CRF, BERT does not constrain the output labels. Thus, the predictions may contain I-ASPECT or I-SENTIMENT without preceding B-ASPECT or B-SENTIMENT. In our case, we found 65 invalid BIO cases when using BERT. Some examples of sentences with invalid token labels are ``...kost(O) nya(I-ASPECT) cukup(B-SENTIMENT) dekat(O)...'' (\textit{...the room is close to...}) and ``...waktu(O) \#\#nya(O) di(O) gant(I-SENTIMENT) \#\#i(I-SENTIMENT) karena(O)...'' (\textit{...need to be changed because...}). We can see that without CRF, BERT can generate the tag sequences quite well. This performance may explain why we only gained a small improvement by adding the CRF layer.

\section{Related work}

\citet{wang2017coupled} summarized several studies on aspect and opinion term extraction. Some of the methods used are association rule mining \citep{hu2004mining}, dependency rule parsers \citep{qiu2011opinion}, conditional random fields (CRF) and hidden Markov model (HMM) \citep{li2010structure, jin2009novel, gojali2016aspect, ekawati2017aspect}, topic modelling \citep{chen2014aspect, zhao2010jointly}, and deep learning \citep{Fernando2019, wang2017coupled, xue2017mtna, xue2018aspect}.

\citet{Fernando2019} combines the idea of coupled multilayer attentions (CMLA) by \citet{wang2017coupled} and double embeddings by \citet{xue2018aspect} on aspect and opinion term extraction on SemEval. The work by \citet{xue2018aspect} itself is an improvement from what their prior work on the same task \citep{xue2017mtna}. Thus, we only included the work by \citet{Fernando2019} because they show that we can get the best result by combining the latest work by \citet{wang2017coupled} and \citet{xue2018aspect}.

In their paper, \citet{devlin2018bert} show that they can achieve state-of-the-art performance not only on sentence-level but also on token-level tasks, such as for named entity recognition (NER). This conclusion motivates us to explore BERT in our study. This way, we do not need to use dependency parsers or any feature engineering. A recent study \citep{yanuar2020aspect} shows that BERT could achieve an overall $F_1$ score of 0.738 for aspect extraction for tourist spot reviews in bahasa Indonesia. However, they did not use CRF in their study.

\citet{souza2019portuguese} proposed a different approach of BERT with CRF for Portuguese NER. Their method disregard the subwords in the CRF layer instead of using auxiliary labels. They found that the Portuguese pretrained models perform better than the multilingual. However, to the best of our knowledge, the pretrained BERT for bahasa Indonesia is not yet available.

End-to-end ABSA with BERT-CRF can be found in \citep{li-etal-2019-exploiting}. While they could achieve $F_1$ scores of 60.78\% and 74.06\% on LAPTOP and REST categories respectively on re-prepared SemEval dataset provided by \citet{li2019unified}, our study focuses in reviews in bahasa Indonesia as we would like to evaluate the multilingual model of BERT.

\section{Conclusions and future work}

Our work shows that pretrained multilingual BERT with adjusted CRF can achieve similar $F_1$ scores to CMLA and double embeddings in aspect and opinion term extraction task with BIO scheme in noisy bahasa Indonesia text. The main advantage is the number of epochs, where we only require 2-4 epochs to fine-tune instead of 200 epochs needed by \citet{Fernando2019}. Moreover, there is no need to produce the word embedding beforehand, making our solution ready for end-to-end settings. For both token and entity level, adding the CRF layer to BERT results in up to 2\% absolute increase in $F_1$ scores on our labels of interest. We also achieved the best $F_1$ scores for classification at the entity level.

In the future, we aim to compare several transformer-based models, such as XLNet \citep{yang2019xlnet}, XLM \citep{lample2019cross}, and RoBERTa \citep{liu2019roberta} when they are trained using multilingual datasets that include text in bahasa Indonesia as well.

\bibliography{emnlp2018}
\bibliographystyle{acl_natbib_nourl}

\end{document}